\documentclass[conference,a4paper]{IEEEtran}
\IEEEoverridecommandlockouts

\usepackage{tcolorbox}
\usepackage[table]{xcolor}
\usepackage{cite}
\usepackage{amsmath,amssymb,amsfonts}
\usepackage{algorithmic}
 \usepackage{verbatim}
\usepackage{graphicx}
\usepackage{textcomp}
\usepackage{xcolor}
\usepackage{url}
\usepackage{verbatim}
\usepackage{booktabs}
\usepackage{multirow}
\usepackage{booktabs}  \usepackage{threeparttable}
\usepackage{tikz}
\usepackage{orcidlink}
\hypersetup{hidelinks}
\usetikzlibrary{arrows.meta,positioning,fit,backgrounds,calc}
\definecolor{ink}{HTML}{1F2733}
\definecolor{mute}{HTML}{5B6675}
\definecolor{genbg}{HTML}{EEF2F7}
\definecolor{selbg}{HTML}{FBF3E4}
\definecolor{misbg}{HTML}{E8F2EE}
\definecolor{llmedge}{HTML}{2F5C94}
\definecolor{ssvedge}{HTML}{B9770C}
\definecolor{gtedge}{HTML}{2B7D67}
\definecolor{ioedge}{HTML}{8A93A3}

\def\BibTeX{{\rm B\kern-.05em{\sc i\kern-.025em b}\kern-.08em
    T\kern-.1667em\lower.7ex\hbox{E}\kern-.125emX}}

\begin{document}

\title{SFE-O: An Ontology for Knowledge-Grounded AI in Supercritical CO$_2$ Biomass Extraction and Fractionation}

\author{\IEEEauthorblockN{Thomas Pallet}
\IEEEauthorblockA{
\textit{Université de Lorraine, CNRS, CRAN} \\
Nancy, France \\
\textit{LIFIA, Facultad de Inform{\'a}tica} \\
\textit{Universidad Nacional de La Plata} \\
La Plata, Argentina \\
\textit{MG-IB} \\
Épinal, France \\
ORCID: 0009-0000-3240-0269
}
\and
\IEEEauthorblockN{Mario Lezoche}
\IEEEauthorblockA{
\textit{Université de Lorraine, CNRS, CRAN} \\
Nancy, France \\
ORCID: 0000-0002-3271-1742
}
\and
\IEEEauthorblockN{Gianpaolo Leone}
\IEEEauthorblockA{
\textit{ENEA, Italian National Agency for New} \\
\textit{Technologies, Energy and Sustainable} \\
\textit{Casaccia Research Centre} \\
Rome, Italy \\
ORCID: 0000-0001-5165-4520
}
\and
\IEEEauthorblockN{Daniele Pizzichini}
\IEEEauthorblockA{
\textit{ENEA, Italian National Agency for New} \\
\textit{Technologies, Energy and Sustainable} \\
\textit{Casaccia Research Centre} \\
Rome, Italy \\
ORCID: 0000-0003-0544-9993
}
\and
\IEEEauthorblockN{Diego Torres}
\IEEEauthorblockA{
\textit{LIFIA, CICPBA-Facultad de Informática} \\
\textit{Universidad Nacional de La Plata} \\
La Plata, Argentina \\
\textit{SIGMA, Depto CyT, UNQ, Argentina} \\
ORCID: 0000-0001-7533-0133
}
\and
\IEEEauthorblockN{Hervé Panetto}
\IEEEauthorblockA{
\textit{Université de Lorraine, CNRS, CRAN} \\
Nancy, France \\
ORCID: 0000-0002-5537-2261
}
\and
\IEEEauthorblockN{Mélanie Glibusic}
\IEEEauthorblockA{
\textit{MG-IB} \\
Épinal, France \\
ORCID: 0009-0003-9802-0164
}

}

\maketitle

\begin{abstract}
Supercritical CO$_2$ extraction (CO$_2$-SFE) offers a sustainable route for recovering high-value compounds from biomass while limiting the use of conventional organic solvents, but its design and interpretation require heterogeneous knowledge spanning biomass properties, operating conditions, mass transfer, and product fractionation, typically fragmented across technical reports, engineering models, external databases, and expert practice. This paper introduces SFE-O, a modular Web Ontology Language (OWL~2) ontology for supercritical CO$_2$ biomass extraction, structured around interconnected Biomass, Process, and Product dimensions and extended with explicit representations of process topology, parameter provenance, external and internal mass-transfer phenomena, and the Constant Extraction Rate (CER) / Falling Extraction Rate (FER) mechanisms. Computational dependencies among process quantities are represented explicitly, without embedding numerical solving in the ontology, so that input, measured, computed, externally retrieved, and fitted parameters remain traceable. SFE-O reuses established semantic resources (ChEBI, QUDT, OWL-Time, PROV-O) and is evaluated through a worked example that instantiates a real pilot-scale extraction batch, a reasoner-based consistency check, and a competency-question SPARQL query. The resulting knowledge model provides a machine-actionable foundation for querying, consistency checking, and mechanistic explanation of SFE processes, and establishes a knowledge layer that could support future ontology-grounded AI reasoning, a direction this paper motivates but does not evaluate.
\end{abstract}

\begin{IEEEkeywords}
Supercritical CO$_2$ extraction, ontology engineering, knowledge representation, semantic interoperability, biomass valorisation, mass-transfer modelling, knowledge-grounded AI.
\end{IEEEkeywords}

\section{Introduction}
\label{sec:intro}

The valorisation of biological resources requires extraction technologies capable of recovering high-value compounds while limiting environmental impact, solvent consumption, and degradation of thermosensitive molecules. In this context, supercritical carbon dioxide extraction (CO$_2$-SFE) has emerged as a particularly relevant technology for processing plant and microbial biomass. Above its critical point, CO$_2$ combines a liquid-like density with gas-like transport properties, while its solvent power can be tuned through pressure and temperature. These characteristics enable selective extraction under comparatively mild operating conditions and make CO$_2$-SFE attractive for applications in the food, nutraceutical, pharmaceutical, and cosmetic sectors \cite{Nozari2025,Yan2024}.

The design and interpretation of an SFE process nevertheless require substantially more knowledge than the specification of extraction pressure and temperature alone. Process behaviour depends simultaneously on the chemical composition and morphology of the biomass, pretreatment conditions, particle-size distribution, extraction-vessel geometry, fixed-bed characteristics, CO$_2$ flow rate, possible co-solvent use, thermophysical fluid properties, molecular descriptors of the target compounds, hydrodynamic conditions, external and internal mass-transfer mechanisms, extraction time, and the configuration of the downstream separation stages. These quantities are mutually dependent and may originate from direct measurements, engineering calculations, external physicochemical databases, mathematical models, or parameter-fitting procedures. At pilot and industrial scales, the problem is further extended by closed-loop CO$_2$ circulation and by sequential separators operated under different pressure and temperature conditions, allowing the extracted material to be fractionated according to the solubility behaviour of its constituents.

The kinetics of SFE also involve mechanistic knowledge that cannot be reduced to a flat collection of process parameters. The Broken-and-Intact Cell (BIC) approach distinguishes an easily accessible solute fraction associated with structures disrupted during biomass pretreatment from a less accessible fraction retained within intact structures \cite{Sovova1994,Sovova2017}. This distinction underlies the interpretation of a fast extraction region dominated by external mass transfer and a subsequent region increasingly governed by internal diffusion. Consequently, quantities such as the broken-cell fraction, Reynolds, Schmidt, and Sherwood numbers, molecular and effective diffusivities, and external mass-transfer coefficients form a network of physical and computational dependencies that is necessary for explaining how process conditions influence extraction behaviour.

While established, process engineering knowledge remains fragmented across literature, technical documents, code, and expert practice, hindering its computational use. To address this, general-purpose ontologies like OntoCAPE provide formal, reusable conceptualizations that support knowledge management and engineering applications \cite{Morbach2007,Morbach2009}. However, a general process ontology does not by itself provide the domain-specific representation required to connect biomass morphology and composition with SFE operating conditions, transport mechanisms, fractional separation, derived quantities, and the provenance of the knowledge used to calculate them.

This paper addresses this problem through \emph{SFE-O}, a modular ontology for the formal representation of supercritical CO$_2$ biomass extraction knowledge. SFE-O structures the domain around three interconnected dimensions, \emph{Biomass}, \emph{Process}, and \emph{Product}, while introducing dedicated representations for process topology, temporal phases, transport phenomena, computational dependencies, and parameter provenance. Rather than representing domain quantities as isolated datatype attributes, the ontology explicitly models their semantic context and derivation relationships, distinguishing values that are provided as inputs, directly measured, computationally derived, retrieved from external resources, or estimated through fitting. The ontology also formalises the relationship between biomass disruption and the Constant Extraction Rate (CER) and Falling Extraction Rate (FER) mechanisms, thereby connecting structural characteristics of the feedstock to external and internal mass-transfer phenomena.

The contribution is therefore not intended to replace mechanistic or numerical SFE models. Instead, SFE-O provides a machine-actionable semantic layer in which such models, their required inputs, their outputs, and their provenance can be explicitly interconnected. This layer is designed to support semantic querying, consistency checking, traceable mechanistic explanations, and interoperability with established knowledge resources. In accordance with this objective, the present work focuses on knowledge formalization and ontology validation and does not expose or evaluate experimental datasets.



\section{Related Work}
\label{sec:related_work}

The development of SFE-O lies at the intersection of three research areas: mechanistic modelling of supercritical-fluid extraction, semantic modelling for chemical and bioprocess engineering, and reusable ontologies for representing chemical entities, quantities, temporal information, and provenance. Existing work provides important components for each of these areas, but does not directly address their integration into a machine-actionable representation of supercritical CO$_2$ biomass extraction.

\subsection{Process and Biomass Ontologies}

Ontologies have been investigated in chemical and process engineering as a means to make engineering knowledge explicit, reusable, and interoperable across heterogeneous software environments. A major contribution in this direction is OntoCAPE, a large-scale formal ontology for Computer-Aided Process Engineering (CAPE) \cite{Morbach2007,Morbach2009}. OntoCAPE provides a modular and layered representation of process-engineering concepts, including plant equipment, process systems, material streams, physical quantities, and mathematical models. Its architecture demonstrates how formal semantics can support knowledge management, process modelling, simulation-related applications, and reuse across different engineering tasks.

The biomass and biorefinery domain has also motivated more specialised semantic models. Trokanas \emph{et al.} proposed BiOnto, an ontology representing biomass types, biomass composition, and biorefining processing technologies \cite{Trokanas2015}. BiOnto was designed to support the identification and comparison of biomass-processing pathways and demonstrated the use of semantic reasoning for applications such as technology screening and industrial symbiosis. Subsequent work has further explored ontology-supported characterisation of biorefinery streams and the use of semantic knowledge to improve decision making in biorefinery value chains \cite{Bussemaker2018,Trokanas2016}.

These approaches establish the relevance of ontology engineering for both process systems and biomass valorisation. However, their modelling scope is broader than the specific mechanisms governing supercritical-fluid extraction. In particular, the representation required for CO$_2$-SFE must connect biomass morphology and chemical composition with extraction-system topology, supercritical-fluid properties, hydrodynamics, molecular transport, extraction kinetics, and fractional separation. It must also distinguish the origin of process quantities and explicitly represent dependencies among parameters used by engineering correlations and mechanistic models.

\subsection{Reusable Semantic Resources}

Rather than reproducing knowledge already formalised in established resources, domain ontologies can improve interoperability by reusing or aligning existing semantic models. Chemical entities are particularly relevant in SFE because extraction behaviour is molecule dependent. The Chemical Entities of Biological Interest ontology (ChEBI) provides a curated ontology and reference vocabulary for chemical entities and their structural and functional classifications \cite{Hastings2016}. Its hierarchy and stable identifiers enable molecules such as fatty acids and carotenoids to be connected to broader chemical classes without reconstructing an independent chemical taxonomy. Functional roles represented in ChEBI are also useful for linking extracted compounds to downstream interpretations such as antioxidant, pigment, or nutritional functions.

For biomass-derived products intended for food-related applications, FoodOn provides an interoperable ontology for food materials, their biological origins, processing, and derived products \cite{Dooley2018}. Such external domain resources complement SFE-specific knowledge by allowing product fractions to be connected to broader application and valorisation contexts.

Engineering knowledge additionally requires a consistent representation of physical quantities and units. The Quantities, Units, Dimensions and Types ontology (QUDT) provides ontologies for quantities, quantity kinds, units, dimensions, and numerical quantity values \cite{QUDT2026}. This distinction is important for SFE knowledge, where quantities such as pressure, temperature, mass flow, density, viscosity, diffusivity, and extraction yield are expressed using heterogeneous units and may also depend on a reference basis such as dry or wet biomass.

Temporal semantics are required because an extraction process comprises ordered stages and time-dependent regions. OWL-Time provides concepts for temporal intervals, durations, instants, and their ordering relations \cite{OWLTime2017}. These constructs can be specialised for representing extraction batches, extraction cycles, transient periods, operating-regime intervals, and kinetic extraction phases.

Finally, the origin of a parameter is essential to its interpretation. A value may be experimentally measured, supplied as an operating set point, computed from other quantities, retrieved from an external physicochemical resource, or estimated through parameter fitting. The W3C PROV Ontology (PROV-O) provides a general provenance model based on entities, activities, agents, and derivation relations \cite{Lebo2013}. Its specialisation therefore enables the derivation chain of an SFE quantity to be represented explicitly rather than treating all numerical values as semantically equivalent.

\subsection{Research Gap and Positioning of SFE-O}

The reviewed approaches provide important building blocks but leave a modelling gap at their intersection. General process ontologies describe engineering systems at a broad level, whereas biomass and biorefinery ontologies primarily address feedstock classification, processing technologies, and transformation pathways. Conversely, chemical and metrological ontologies provide reusable semantics for individual information dimensions but do not describe how these dimensions interact within an SFE extraction mechanism, as summarised in Table~\ref{tab:related_work_comparison}.

\begin{table}[htbp]
\caption{Positioning of SFE-O relative to related ontologies}
\label{tab:related_work_comparison}
\centering
\footnotesize
\begin{tabular}{p{0.16\columnwidth} p{0.34\columnwidth} p{0.40\columnwidth}}
\hline
\textbf{Ontology} & \textbf{Scope} & \textbf{Gap addressed by SFE-O} \\
\hline
OntoCAPE \cite{Morbach2007,Morbach2009} & General CAPE process systems, equipment, streams & No SFE-specific mass-transfer, CER/FER, or fractionation semantics \\
BiOnto \cite{Trokanas2015,Trokanas2016,Bussemaker2018} & Biomass/biorefinery feedstock and conversion pathways & No extraction-process topology or kinetic mechanisms \\
ChEBI, QUDT, OWL-Time, PROV-O \cite{Hastings2016,QUDT2026,OWLTime2017,Lebo2013} & Chemical identity, units, time, provenance (reused by SFE-O) & Provide reusable semantics but no SFE process model \\
\textbf{SFE-O (this paper)} & SFE-specific: biomass--process--product, BIC/CER-FER, parameter provenance & --- \\
\hline
\end{tabular}
\end{table}

To our knowledge, no existing ontology specifically formalises the supercritical CO$_2$ extraction domain, combining Broken-and-Intact Cell (BIC) extraction kinetics with explicit parameter provenance; this motivates the SFE-specific contribution below.

SFE-O addresses this gap rather than competing with the aforementioned ontologies; instead, it specialises the extraction domain while reusing their semantics wherever appropriate, providing the formal structure required for traceable reasoning and subsequent knowledge-grounded AI applications.

\section{Knowledge Elicitation and Ontology Requirements}
\label{sec:knowledge_elicitation}

The construction of SFE-O followed an iterative knowledge-engineering process in which domain knowledge was progressively extracted, structured, reconciled, and formalised. The methodology is a lightweight adaptation of the Ontology Development 101 iterative cycle \cite{NoyMcGuinness2001}, combined with the competency-question method of Gr\"uninger and Fox \cite{GruningerFox1995}, applied here through five steps. In the present work, the objective was not to reproduce experimental datasets semantically, but to identify the concepts, relations, constraints, and computational dependencies required to describe and reason about a supercritical CO$_2$ extraction process.

\subsection{Knowledge Sources and Elicitation Process}

The domain conceptualization was derived from six heterogeneous knowledge artefacts produced during the analysis of CO$_2$-SFE processes. They comprised two technical reports describing bench- and pilot-scale extraction, a structured parameter-reference document, a computational spreadsheet relating process quantities, a set of conceptual hierarchy diagrams, and an expert-authored descriptive model of the extraction mechanisms. Their complementary nature was important because no individual source contained all the knowledge required by the ontology.

The bench-scale material provided the terminology associated with biomass characterisation, pretreatment, extraction equipment, target chemical compounds, operating conditions, co-solvent use, extraction time, and analytical characterisation. The pilot-scale documentation extended this description towards process topology by introducing closed-loop CO$_2$ circulation, extraction cycles, equipment configuration, sequential separation stages, and the possibility of operating separators under distinct pressure and temperature conditions. The structured parameter material established correspondences among biomass properties, chemical compounds, process variables, units of measurement, and chemical identifiers. The computational representation exposed dependencies among quantities, whereas the conceptual diagrams organised these quantities into domain hierarchies and distinguished different origins of parameter values. Finally, the descriptive knowledge source provided explicit definitions of the relevant entities, extraction phases, physical mechanisms, equations, and domain constraints.

Knowledge elicitation was performed by iteratively moving from terminology to conceptual dependencies. First, recurrent domain terms were identified and grouped according to their function in the extraction process (\emph{terminology elicitation}). Second, relationships among these terms were extracted, including structural, temporal, chemical, process, and computational relationships (\emph{relation extraction}). Third, mathematical expressions and engineering correlations were analysed to identify which quantities act as inputs and outputs of calculations. Fourth, domain constraints and modelling assumptions were separated from taxonomic knowledge (\emph{constraint separation}). Finally, the resulting conceptualization was reviewed with domain experts (\emph{expert review}) and reorganised into reusable ontology patterns.

A major outcome of this process was the identification of three primary and causally connected domain perspectives:

\begin{equation}
\mathrm{Biomass} \longrightarrow \mathrm{Process} \longrightarrow \mathrm{Product}.
\label{eq:bpp}
\end{equation}

The \emph{Biomass} perspective describes the material entering the extraction process, including its type, chemical composition, morphology, physical characteristics, and pretreatment state. The \emph{Process} perspective represents the extraction system, its geometry, equipment, operating parameters, fluid properties, temporal phases, hydrodynamics, and mass-transfer mechanisms. The \emph{Product} perspective describes the extract and its fractions, their chemical composition, and quantities such as extracted amount, yield, recovery, efficiency, and concentration. Equation~\ref{eq:bpp} does not represent a purely sequential taxonomy: information from the three perspectives is linked through causal and computational dependencies.

\subsection{Ontology Requirements}

The elicitation process resulted in eight principal modelling requirements, summarized in Table~\ref{tab:requirements}.

\begin{table}[t]
\caption{Main modelling requirements of SFE-O}
\label{tab:requirements}
\centering
\footnotesize
\begin{tabular}{p{0.08\columnwidth} p{0.82\columnwidth}}
\hline
\textbf{Req.} & \textbf{Description} \\
\hline
R1 & Represent Biomass, Process, and Product as interconnected but modular domain perspectives. \\
R2 & Reuse external semantic resources for chemical entities, quantities, units, time, and provenance whenever appropriate. \\
R3 & Represent physical and chemical quantities independently from their numerical values, units, and reference bases. \\
R4 & Distinguish intrinsic entity types from contextual roles, e.g., a chemical entity playing the role of solute or co-solvent in a process. \\
R5 & Distinguish input, measured, computed, externally retrieved, and fitted parameter values and preserve their provenance. \\
R6 & Represent dependencies among quantities and mechanistic models without using OWL as a numerical solver. \\
R7 & Represent extraction batches, cycles, stages, and an arbitrary ordered sequence of separation stages rather than a plant-specific fixed topology. \\
R8 & Represent domain constraints and preserve distinctions among missing, below-detection-limit, and actual zero-valued information. \\
\hline
\end{tabular}
\end{table}

The remaining requirements are briefly motivated before the two most structurally demanding ones, R3 and R5--R6, are discussed in detail. R1 and R2 shape the overall architecture: R1 reflects the three causally linked perspectives identified during elicitation (Eq.~\ref{eq:bpp}), realised through the separate but interconnected \texttt{sfeo-biomass}, \texttt{sfeo-process}, and \texttt{sfeo-product} modules (Section~\ref{sec:ontology_architecture}), while R2 avoids re-formalizing chemical identity, units, time, and provenance semantics already covered by ChEBI, QUDT, OWL-Time, and PROV-O (Section~\ref{sec:related_work}). R4 and R7 keep the model generic rather than plant-specific: R4 treats roles such as solute or co-solvent as contextual and process-dependent rather than intrinsic to a chemical entity, since the same compound can play different roles across batches, while R7 generalizes separation topology to an ordered sequence of stages (Eq.~\ref{eq:separator_sequence}) instead of fixed, plant-specific separator classes. Finally, R8 preserves the distinction between a missing value, a below-detection-limit measurement, and an actual measured zero, a distinction that matters wherever all three occur in product-composition data (Section~\ref{sec:evaluation}).

Requirement R3 was particularly relevant because the source material combines quantities expressed using different units and reference bases. A chemical concentration, for example, is not represented merely as a molecule-specific datatype property; it is modelled as a contextualized quantity concerning a chemical entity and a material, associated with a numerical value, unit, and, where necessary, a reference basis such as dry or wet biomass. This design prevents the ontology schema from growing whenever a new chemical compound is introduced.

Requirement R5 emerged from the observation that process knowledge combines quantities having different epistemic origins. An extraction pressure may be introduced as an operating set point, biomass composition may result from measurement, Reynolds or Sherwood numbers may be calculated, CO$_2$ thermophysical properties may be obtained from an external computational resource, while parameters such as the broken-cell fraction or effective diffusivity may be estimated by fitting. Making these distinctions explicit is necessary for traceability and for subsequent explanation of how a result has been obtained.

Requirement R6 concerns the mechanistic structure of the extraction process. The elicited knowledge identifies dependencies linking biomass morphology, process geometry, fluid properties, molecular descriptors, hydrodynamics, and mass transfer. For instance, Reynolds and Schmidt numbers contribute to the determination of the Sherwood number, which is subsequently involved in the determination of the external mass-transfer coefficient. These relations must therefore be represented as explicit computational dependencies while leaving their numerical evaluation to appropriate external software or mathematical services.

The same requirement applies to the distinction between the Constant Extraction Rate (CER) and Falling Extraction Rate (FER) regions. The broken-cell fraction $g$ identifies the portion of the biomass whose solute is immediately accessible after pretreatment and is therefore associated with the faster, externally controlled extraction mechanism. Conversely, the remaining intact fraction $(1-g)$ introduces internal mass-transfer resistance and contributes to the diffusion-controlled extraction mechanism. Preserving this mechanistic distinction was considered essential for supporting explanations that connect pretreatment and biomass morphology to extraction behaviour.

\subsection{Competency Questions}

The requirements were complemented by competency questions (CQs) expressing representative information needs that SFE-O should support. A subset used to guide the present ontology is reported below:

\begin{itemize}
    \item \textbf{CQ1:} Which biomass characteristics and chemical entities are relevant to a given supercritical CO$_2$ extraction process?
    \item \textbf{CQ2:} Which geometric and operating quantities characterize a given extraction batch and its process stages?
    \item \textbf{CQ3:} Which quantities are required to derive $Re$, $Sc$, $Sh$, and $k_f$, and what is the provenance of each required quantity?
    \item \textbf{CQ4:} How is the broken-cell fraction related to the portions of a target compound associated with the CER and FER mechanisms?
    \item \textbf{CQ5:} Which separation stage generates a given product fraction, and under which pressure and temperature conditions?
    \item \textbf{CQ6:} Which chemical entities or chemical classes characterize a product fraction and which functional roles can be associated with them?
    \item \textbf{CQ7:} Was a process quantity supplied, measured, calculated, retrieved from an external resource, or estimated through fitting?
    \item \textbf{CQ8:} Are the represented quantities and relationships compatible with the domain constraints encoded for the extraction process?
\end{itemize}

These competency questions define the boundary between domain information that merely occurs in the source material and knowledge that must be explicitly accessible through the ontology. They consequently guided the modular architecture, reuse decisions, properties, and constraint representations presented in the following section.

\section{SFE-O Architecture and Semantic Reuse}
\label{sec:ontology_architecture}

\begin{figure*}[t]
    \centering
    \includegraphics[width=0.65\textwidth]{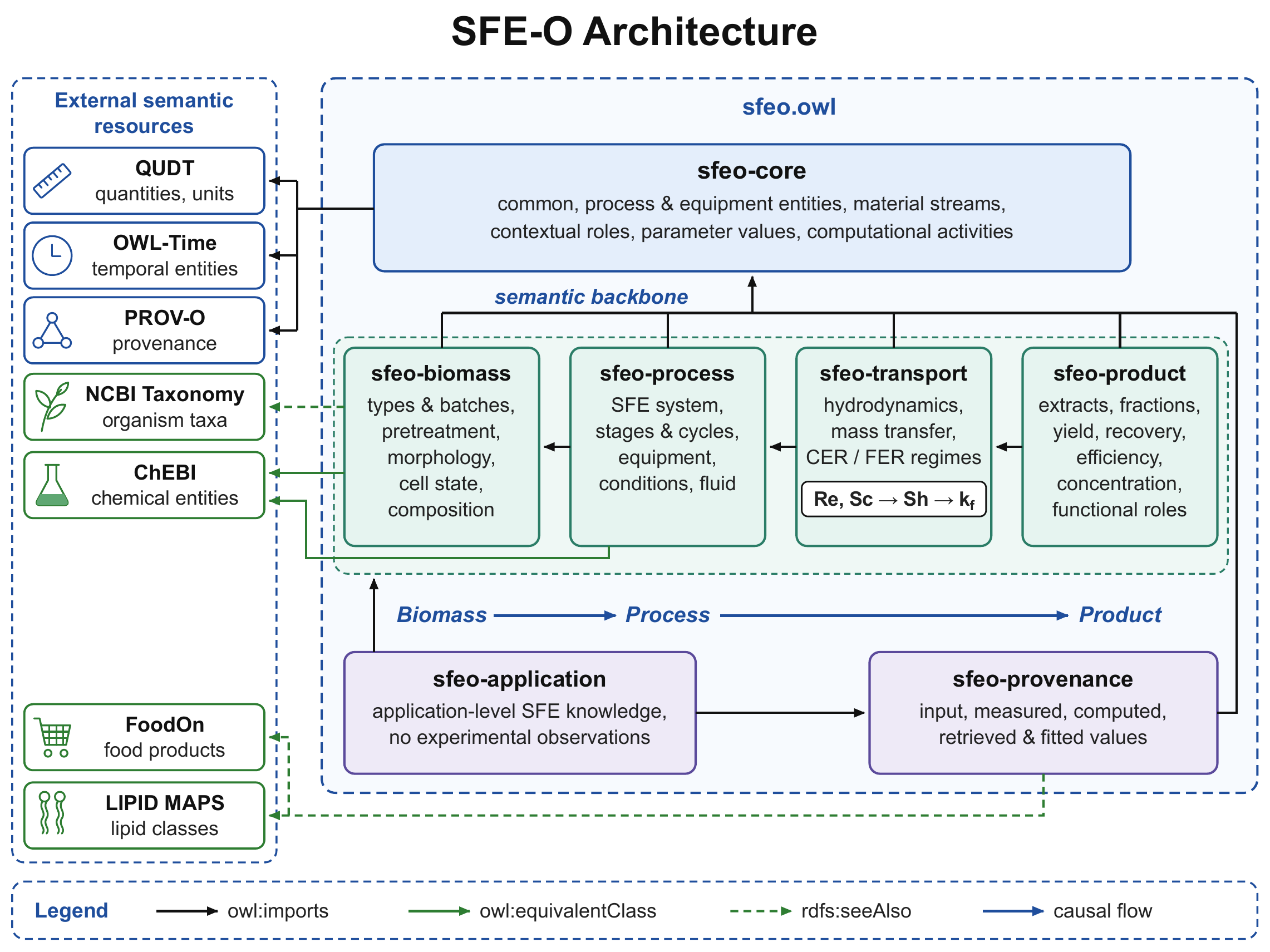}
    \caption{Modular architecture of SFE-O and its relationships with reused or aligned external semantic resources. 
    }
    \label{fig:sfeo_architecture}
\end{figure*}

SFE-O is implemented as a modular OWL~2 ontology whose organisation reflects the separation of concerns identified during knowledge elicitation. The SFE-O v0.1.0 ontology described in this paper is openly available on Zenodo \cite{SFEO2026} and is organised into seven local modules (\emph{Core}, \emph{Biomass}, \emph{Process}, \emph{Transport}, \emph{Product}, \emph{Provenance}, and \emph{Application}) together with an aggregate ontology importing the complete model. Fig.~\ref{fig:sfeo_architecture} summarizes this architecture and its connections with external semantic resources.

\subsection{Modular Architecture}

The \texttt{sfeo-core} module provides the common semantic backbone, including material and process entities, equipment, material streams, contextual roles, parameter values, and activities responsible for producing or deriving knowledge. The \texttt{sfeo-biomass} module specialises this backbone with biomass types and batches, pretreatment operations, morphology, physical characteristics, chemical composition, and the distinction between broken and intact structural units required by the Broken-and-Intact Cell representation.

The \texttt{sfeo-process} module represents extraction systems, batches, cycles, equipment, operating conditions, geometry, fluid properties, and temporal process stages. An \texttt{ExtractionBatch} denotes the complete extraction operation performed on a defined biomass load, whereas an \texttt{ExtractionCycle} represents one circulation of CO$_2$ through the relevant functional stages. This distinction supports both single-pass laboratory systems and closed-loop pilot or industrial configurations.

Separation topology is intentionally generalized. Rather than defining plant-specific classes such as \texttt{SE1}, \texttt{SE2}, and \texttt{SE3}, an extraction cycle may contain an ordered sequence of \texttt{SeparationStage} individuals,

\begin{equation}
S_1 \prec S_2 \prec \dots \prec S_n ,
\label{eq:separator_sequence}
\end{equation}

each associated with its equipment, pressure, temperature, and generated \texttt{ProductFraction}, so a three-separator pilot plant is one instantiation, not a structural limit.

The \texttt{sfeo-transport} module represents hydrodynamic and mass-transfer knowledge, including molecular and effective diffusivity, Reynolds, Schmidt, and Sherwood numbers, the external mass-transfer coefficient $k_f$, and the CER/FER mechanisms. The \texttt{sfeo-product} module describes total extracts and separated fractions together with quantities such as extracted amount, yield, recovery, efficiency, concentration, and functional roles. Finally, \texttt{sfeo-provenance} distinguishes input, measured, computed, externally retrieved, and fitted parameter values, while \texttt{sfeo-application} contains application-level SFE knowledge without representing experimental observations.

\subsection{Semantic Modelling Patterns}

Three modelling patterns are central to SFE-O.

First, chemical composition is represented through a reified relation between a material, a chemical entity, and a quantity value rather than through molecule-specific datatype properties. Conceptually, a composition statement is represented as

\begin{equation}
\langle M,C,Q,R \rangle ,
\label{eq:composition_pattern}
\end{equation}

where $M$ is a biomass or product material, $C$ the chemical entity, $Q$ the associated quantity, and $R$ an optional reference basis such as dry or wet mass (Eq.~\ref{eq:composition_pattern}). New compounds can therefore be introduced without modifying the ontology schema.

Second, SFE-O distinguishes intrinsic entity types from contextual process roles: a chemical entity is not intrinsically a \emph{solute}, but may play a \texttt{SoluteRole} in a given extraction context, and the same pattern applies to solvent and co-solvent participation, since these roles depend on the process configuration and operating conditions.

Third, computational dependencies are explicitly represented without requiring OWL to execute numerical models. A calculation is connected to its required quantities and generated output,

\begin{equation}
\{Q_1,\ldots,Q_n\}
\xrightarrow{\mathrm{Calculation}}
Q_{\mathrm{out}} .
\label{eq:calculation_pattern}
\end{equation}

For example, this pattern (Eq.~\ref{eq:calculation_pattern}) represents that the Sherwood number depends on Reynolds and Schmidt numbers and that the external mass-transfer coefficient subsequently depends on Sherwood number, solute diffusivity, and particle size. Numerical evaluation remains delegated to appropriate engineering software, while the ontology preserves the semantic dependency and provenance chain.

\subsection{Reuse of External Semantic Resources}

SFE-O adopts the selective reuse strategy already motivated in Section~\ref{sec:related_work} (QUDT \cite{QUDT2026}, OWL-Time \cite{OWLTime2017}, PROV-O \cite{Lebo2013}), illustrated on the left side of Fig.~\ref{fig:sfeo_architecture}: quantities and units, temporal concepts, and the provenance of measurement, calculation, external-lookup, and fitting activities are each delegated to the corresponding external ontology rather than re-modelled locally. Chemical entities are likewise selectively aligned with ChEBI \cite{Hastings2016} rather than reconstructed through a local taxonomy, so that compounds such as carotenoids and fatty acids retain stable, externally maintained identifiers while remaining associated with SFE-specific process roles. Other resources serve only as interoperability or extension points, not full imports: NCBI Taxonomy \cite{Schoch2020} for biological-source identifiers, and FoodOn/LIPID MAPS as possible alignments for food-product and lipid-specific semantics.

This architecture keeps SFE-O focused on knowledge specific to supercritical CO$_2$ extraction while delegating chemical identity, units, time, and provenance to established resources.

\section{Mechanistic Knowledge and Constraint Formalization}
\label{sec:mechanistic}

A distinctive feature of SFE-O is the explicit representation of mechanistic relationships connecting biomass structure, operating conditions, mass transfer, and product formation.

\subsection{CER/FER Representation}

Following the BIC interpretation \cite{Sovova1994,Sovova2017}, let $m_s$ be loaded biomass mass, $X_0(i)$ the initial concentration of compound $i$, and $g$ the broken-cell fraction. SFE-O represents the initial, broken-cell, and intact-cell amounts as

\begin{equation}
\begin{aligned}
C_0(i)&=m_sX_0(i),\\
C_{0,BC}(i)&=g\,m_sX_0(i),\quad
C_{0,IC}(i)=(1-g)m_sX_0(i).
\end{aligned}
\label{eq:bic_partition}
\end{equation}

The broken fraction (Eq.~\eqref{eq:bic_partition}) is associated with the Constant Extraction Rate (CER) region, predominantly governed by external mass transfer, whereas the intact fraction contributes to the Falling Extraction Rate (FER) region, increasingly controlled by internal diffusion. The corresponding mass-balance constraints are

\begin{equation}
\begin{aligned}
M_{\mathrm{CER}}&\le g m_sX_0(i),\quad
M_{\mathrm{FER}}\le(1-g)m_sX_0(i),\\
M_{\mathrm{TOT}}&=M_{\mathrm{CER}}+M_{\mathrm{FER}}\le m_sX_0(i).
\end{aligned}
\label{eq:cerfer_constraints}
\end{equation}

Product indicators can then be linked to the same mechanistic model (Eq.~\eqref{eq:cerfer_constraints}), e.g., $EFF=M_{\mathrm{TOT}}/C_0(i)$ and $CONC=M_{\mathrm{TOT}}/EA$, where $EA$ is total extracted material.

\subsection{Hydrodynamic and Mass-Transfer Dependencies}

Extraction pressure and temperature determine supercritical-CO$_2$ properties such as density $\rho$ and viscosity $\mu$; biomass morphology contributes particle size $d_p$; and molecular identity contributes diffusivity $D(i)$. With flow and bed characteristics, these quantities determine $Re$ and $Sc$, followed by

\begin{equation}
Sh=2+1.1Re^{0.6}Sc^{0.33},\qquad
\{Sh,D(i),d_p\}\rightarrow k_f(i).
\label{eq:transport_chain}
\end{equation}

Each derivation is a \texttt{CalculationActivity}, with calibration and external lookups typed as \texttt{ParameterFittingActivity} and \texttt{ExternalLookupActivity}. Effective diffusivity $D_{\mathrm{EFF}}(i)$, governing internal transport, depends on porosity, tortuosity, and microstructure in a biomass-specific way for which the elicited sources give no single universal equation; SFE-O therefore records the dependency without imposing one.

\subsection{Constraints}

OWL axioms express logical conditions such as disjointness of broken/intact structures and CER/FER phases, while arithmetic bounds, parameter ranges, and derived-quantity relations are kept as machine-readable constraint expressions, not evaluated by the OWL reasoner itself. \emph{Not analysed}, \emph{below detection limit}, and a measured zero remain three distinct states, preserving semantic correctness while leaving numerical constraint execution to future validation or engineering services.

\section{Knowledge-Grounded AI Perspective}
\label{sec:ai}

SFE-O is also intended as a grounding layer for future AI-assisted SFE interpretation using a Large Language Model (LLM); no LLM system is built or evaluated in this paper. Work on LLM--knowledge-graph integration highlights the value of combining generative language capabilities with explicit machine-interpretable knowledge \cite{Khorashadizadeh2024}. An LLM could translate a natural-language question into a semantic query over SFE-O and use retrieved facts, mechanistic dependencies, and provenance as grounded context. For example, a question about $k_f$ can expose the path from operating/biomass properties through $Re$, $Sc$, and $Sh$ (Eq.~\eqref{eq:transport_chain}) together with the provenance of the required quantities.

This separates semantic validity from language generation: SFE-O would define admissible concepts, relations, and dependency chains; numerical services would execute calculations; and the LLM would provide interaction and orchestration. The present contribution does not evaluate such an LLM implementation, but establishes the formal layer required for subsequent ontology-grounded question answering and explainable process interpretation.

\section{Evaluation and Discussion}
\label{sec:evaluation}

Evaluation concerns SFE-O as a knowledge representation rather than a predictive system. First, the modular OWL~2 files were checked for RDF/XML parsing and import-structure correctness. Second, the ontology was assessed against the requirements and CQs of Section~\ref{sec:knowledge_elicitation}. It represents biomass and chemical characteristics, process geometry and operating quantities, the $Re$--$Sc$--$Sh$--$k_f$ dependency chain, CER/FER relations, ordered separation stages and product fractions, and the provenance of input, measured, computed, external, and fitted quantities.

Third, CQ1--CQ8 are demonstrated through a worked example. The ontology is a schema with no individuals; one real pilot-scale extraction batch was instantiated using operating data for \emph{Dunaliella salina} (350~bar, 65\textdegree{}C, 0.31~kg/min CO$_2$, 120~min) and its three-separator cascade (SE1/SE2/SE3, with SE2 at 56~bar/25\textdegree{}C and SE3 at 54~bar/25\textdegree{}C), asserting individuals such as \texttt{DS04\_SE2} (a \texttt{SeparationStage}) and \texttt{DS04\_Fraction\_SE2} (the \texttt{ProductFraction} it generates), with a measured $\beta$-carotene concentration (7.08~mg/g, 20.8\% recovery).

The instantiated ontology was classified with the HermiT Description Logic (DL) reasoner and confirmed consistent, satisfying all OWL axioms including the qualified-cardinality restriction on \texttt{generatedBySeparationStage}. The competency question CQ5 (``which separation stage generates a given product fraction, under which pressure and temperature'') was then answered with an executed SPARQL query against the resulting graph:

{\footnotesize
\begin{verbatim}
SELECT ?fraction ?stage ?p ?t WHERE {
 ?fraction sfeo-product:
   generatedBySeparationStage ?stage .
 ?stage sfeo-process:hasOperatingQuantity
   ?pq, ?tq .
 ?pq a sfeo-process:SeparatorPressure ;
     qudt:numericValue ?p .
 ?tq a sfeo-process:SeparatorTemperature ;
     qudt:numericValue ?t . }
\end{verbatim}
}


matching the source value exactly. This single worked example shows CQ1--CQ8 to be expressible and queryable on real data; it does not constitute a performance benchmark.

Fourth, the conceptual model was iteratively reviewed against the elicited engineering knowledge. General separator stages rather than fixed \texttt{SE1--SE3}, reified composition statements, and contextual \emph{solute} roles avoid over-specialisation to one plant or compound set. The main limitations concern execution rather than representation: numerical dependencies are not solved within OWL; constraints are not yet executable Shapes Constraint Language (SHACL) shapes; external thermophysical, descriptor, diffusivity, and fitting services remain integration points; and effective diffusivity requires biomass-specific assumptions. A few data-level inconsistencies also remain open for expert validation: a conflicting interfacial-area formula ($a=6/d_p$ vs.\ $a=d_p/6$) across source documents, a bed-volume expression possibly missing a geometric factor, and one ChEBI identifier shared by two distinct fatty acids; these are flagged explicitly within the ontology rather than silently resolved.

The paper also deliberately excludes broader experimental observations beyond the single worked example above. This establishes a transparent semantic layer before coupling SFE-O with data, numerical models, and generative AI. Future evaluation should therefore instantiate additional batches and competency questions beyond CQ5, test executable constraints, and determine whether ontology-grounded interaction improves traceability, consistency, explanation, and process-space exploration compared with unstructured AI interaction.

\section{Conclusion}
\label{sec:conclusion}

SFE-O provides a modular semantic representation of supercritical CO$_2$ biomass extraction, connecting biomass characteristics, process topology, transport phenomena, separation, products, and provenance. Its central contribution is the explicit formalization of BIC/CER--FER mechanisms and computational dependencies while reusing ChEBI, QUDT, OWL-Time, and PROV-O, demonstrated on an instantiated extraction batch with a HermiT-checked full OWL~2~DL consistency result and an executed competency-question SPARQL query. Future work will add executable constraint validation, numerical/physicochemical services, experimental instantiation, and evaluation of ontology-grounded AI for process interpretation and optimisation.

\section*{Acknowledgment}

Generative AI (Anthropic Claude) was used solely for grammar checking and for rendering Fig.~\ref{fig:sfeo_architecture} to the authors' specifications; the resulting diagram was verified against the ontology's OWL files. All technical content and conclusions remain the authors' own.

\bibliographystyle{IEEEtran}
\bibliography{SFE-O_References_8pages}

\end{document}